# An ensemble framework for explainable geospatial machine learning models

Lingbo Liu [a,b]

[a] *Center for Geographic Analysis, Harvard University, Cambridge, MA 02138, USA*
[b] *School of Urban Design, Wuhan University, Wuhan, Hubei 430072, China*

A B S T R A C T

Analyzing spatially varying effects is pivotal in geographic analysis. However, accurately capturing and interpreting this variability is challenging due to the increasing complexity and non-linearity of geospatial data. Recent advancements in integrating Geographically Weighted (GW) models with artificial intelligence (AI) methodologies offer novel approaches. However, these methods often focus on single algorithms and emphasize prediction over interpretability. The recent GeoShapley method integrates machine learning (ML) with Shapley values to explain the contribution of geographical features, advancing the combination of geospatial ML and explainable AI (XAI). Yet, it lacks exploration of the nonlinear interactions between geographical features and explanatory variables. Herein, an ensemble framework is proposed to merge local spatial weighting scheme with XAI and ML technologies to bridge this gap. Through tests on synthetic datasets and comparisons with GWR, MGWR, and GeoShapley, this framework is verified to enhance interpretability and predictive accuracy by elucidating spatial variability. Reproducibility is explored through the comparison of spatial weighting schemes and various ML models, emphasizing the necessity of model reproducibility to address model and parameter uncertainty. This framework works in both geographic regression and classification, offering a novel approach to understanding complex spatial phenomena.

## 1. Introduction

The relationships between phenomenon can vary significantly across different spatial or geographical contexts, manifesting in events such as the disparate impacts of pandemics (Hammer, 2021); the dynamics of poverty distribution (Chaves, 2015), housing prices fluctuations (Liu, 2022), etc. Optimizing spatial analysis methods is crucial for exploring these diverse issues, as it enhances predictions accuracy, model interpretability, and the effectiveness of spatial decisions or interventions (Brunsdon et al., 1998). Nonetheless, the inherent complexity of spatial data and the potential for nonlinear relationships pose challenges to enhancing interpretability through traditional spatial analysis techniques (De Sabbata, et al., 2023).

For models analyzing spatially varying effects, such as spatial filtering models (Oshan and Fotheringham, 2018; Griffth, 2003; Gorr and Olligschlaeger, 1994) and spatial Bayes models (Oshan, 2022); Geographically Weighted Regression (GWR) and Multiscale Geographically Weighted Regression (MGWR) stand out for their application of local spatial weighting schemes, which capture spatial features more accurately (Murakami, 2020; Fotheringham et al., 2017). These linear regression-based approaches, however, encounter significant hurdles in decoding complex spatial phenomena (Fig. 1). Various Geographically Weighted (GW) models have been developed to tackle issues such as multicollinearity (Wheeler, 2009; Comber and Harris, 2018) and to extend the utility of GW models to classification tasks (Atkinson, 2003; Paez, 2006; Brunsdon et al., 2007; Jiang, 2012). The evolution of artificial intelligence (AI) methodologies, including Artificial Neural Networks (ANN) (Du, 2020), Graph Neural Networks (GNN) (Zhu, 2020; Zhu, 2022), and Convolutional Neural Networks (CNN) (Dai, 2022), has introduced novel ways to mitigate uncertainties around spatial proximity and weighting kernels in GW models. Despite these advancements in combining geospatial models with AI, challenges remain in addressing nonlinear correlations and deciphering underlying spatial mechanisms.

The growing interest in Geospatial AI (GeoAI) and the integration of explicit geographical relationships with AI models herald an important trend towards improving predictions regarding nonlinear relationships and complex spatial phenomena (Mai, 2022; Li, 2021; Gao, 2021). This trend bifurcates into two categories (Li and Hsu, 2022; Liu and Biljecki, 2022). One integrates spatial weighting scheme into AI models, such as GW Support Vector Regression (Yang, 2023), GWANN (Hagenauer and Helbich, 2022), and GW Extreme Learning Machine (Deng et al., 2017), GW Random Forest (Georganos, 2021; Grekousis, 2022), ANN for Geographically and temporally weighted regression(GTWR) (Feng, 2021; Wu, 2021). These models focus predominantly on prediction with less emphasis on interpretability. Moreover, these models often concentrate on a single algorithm and are implemented in various programming languages, hampering their replicability and the ability to perform stable comparisons (Kedron et al., 2023). The other category involves explainable AI (XAI) algorithms, such as Local Interpretable Model-agnostic Explanations (LIME) (Ribeiro et al., 2016); SHapley Additive exPlanations (SHAP) (Lundberg and Lee, 2017), or Feature Importance in tree models (Breiman, 2001), which offer valuable tools for demystifying the "black box" of machine learning (ML) algorithms, enhancing geographic analysis from an interpretability perspective (Masrur, 2022; Liu et al., 2023). Recently, an ensemble method named GeoShapley was proposal to offer a universal analytical method for ML methods, providing spatially varying coefficients, particularly in assessing the contribution of geographical coordinates in ML models by Shapley (Li, 2022). However, the direct interpretation results from SHAP or Shapley values often contain substantial noise, necessitating spatial interpolation through GWR to refine the analysis. This requirement presents a challenge for the direct application of GeoShapley, indicating a need for improved methodologies that can leverage the strengths of XAI. Additionally, there is a concern regarding the lack of exploration of nonlinear interactions between geographical features and explanatory variables.

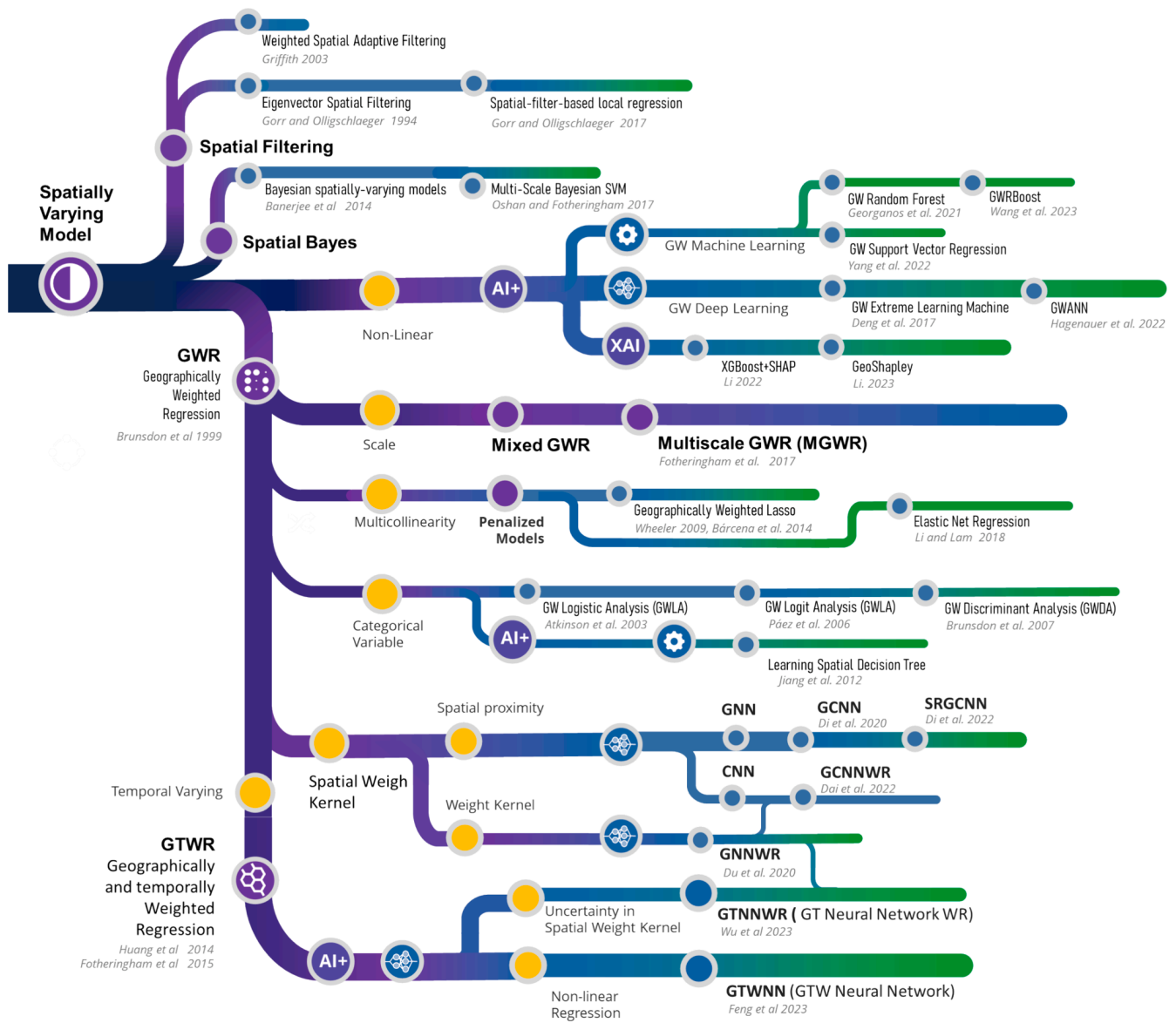

**Fig. 1.** The evolution tree of models for spatially varying effect.

An integrated, explainable geospatial machine learning ensemble framework is thus proposed building upon the foundational assumptions of geographic weighting while incorporating XAI technologies. This framework leverages the spatial weighting feature capture of GWR and the powerful functionality of AI models, bolstering model interpretability through XAI and addressing the shortcomings of conventional approaches. The multi-model testing on synthetic datasets illustrates the framework's enhanced capability to accurately capture and elucidate spatial variability, providing robust support for decision-making and predictive modeling across various spatial studies. This research underscores the potential of this ensemble framework in augmenting prediction accuracy and applicability and presents a versatile approach capable of addressing a wide spectrum of spatial variability challenges, making it readily applicable to practical spatial analysis scenarios.

This study makes three specific contributions to the field of spatially varying models. First, it emphasizes that the overall predictive accuracy of geographic models does not necessarily indicate the accuracy of spatially varying coefficients, necessitating comparisons across multiple models to address this uncertainty. Second, it extends the application of local geographic weighting methods from linear regression to more diverse machine learning models, effectively handling complex and nonlinear relationships. Third, the integration of partial dependency estimation methods aids in explaining the nonlinear relationships between explanatory variables and dependent variables within a geographic context. The framework's ability to handle diverse spatial phenomena makes it a valuable tool for geographers seeking to apply advanced analytical methods to real-world spatial problems.

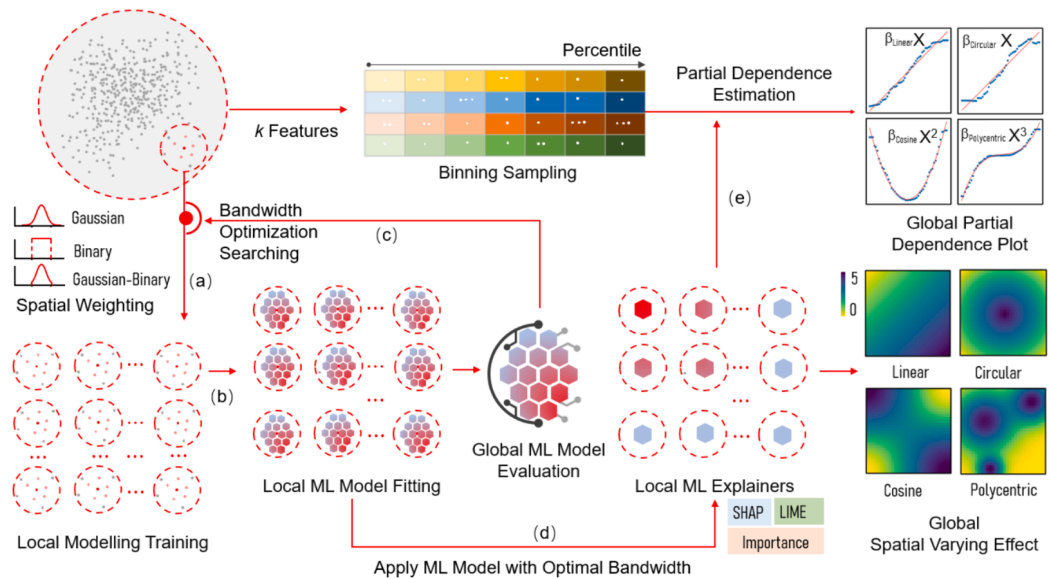

**Fig. 2.** Workflow of the Explainable Geospatial Machine Learning (XGeoML) Model. (a) Initial Bandwidth and Weighting Scheme: Select default bandwidth and appropriate spatial weighting scheme; (b) Local Model Construction & Evaluation: Construct local models for each point; use leave-one-out cross-validation for machine learning training, prediction, and evaluation; (c) Optimal Bandwidth Search: Iterate over bandwidths to find the optimal setting for maximum model performance; (d) Integration & Explanation: Combine the optimal bandwidth with the machine learning model; apply SHAP, LIME, and Feature Importance for interpretable spatial variability coefficients. (e) Partial Dependency Estimation: Use percentile binning sampling on *k* features for partial dependency estimations with the trained model.

## 2. Data and methodology

### 2.1. Data

A systematic approach was employed to generate a synthetic dataset designed to simulate complex spatial relationships, leveraging a 30×30 spatial grid as the foundational structure. Each point within this grid was assigned four independent random variables, each drawn from a standard normal distribution **N** (0, 1). Additionally, an error term (ϵ) was introduced to each point, originating from a normal distribution with a mean of zero and a standard deviation of 0.5, to emulate observational errors commonly encountered in empirical data.

To incorporate spatial variability into the dataset, four distinct spatial gradient patterns were devised: a linear gradient, set at a 45-degree angle to create a continuous variable across the 0 to 1 range ($\beta_{linear}$); a circular gradient centered within the grid to simulate radially diminishing effects ($\beta_{circular}$); a cosine gradient to introduce regular spatial fluctuations($\beta_{cosine}$); and a kernel density gradient, employing Gaussian kernels placed at strategic locations to simulate localized clustering phenomena($\beta_{polycentric}$). These gradients collectively aim to mimic the diverse spatial effects observed in real-world data.

To ensure model training stability and facilitate model validation, all spatial gradient variables were normalized to a uniform range of 0 to 5. This normalization was critical for maintaining variable consistency across the dataset. Building upon the simulated spatial variability, a linear response variable ($Y_0$) was constructed as per Eq. (1). This was achieved through a linear combination of the spatial gradient variables and the random variables, incorporating the error term to introduce variability. Furthermore, a nonlinear response variable (*Y*) was formulated to capture more complex spatial dynamics. This variable integrates the spatial gradient variables and the random variables in a similar fashion but introduces nonlinearity through the transformation of $X_3$ and $X_4$, which are squared and cubed, respectively, as detailed in Eq. (2). These transformations were specifically chosen to enhance the representation of complex spatial relationships within the model.

$$Y_0 = \beta_{linear}X_1 + \beta_{circular}X_2 + \beta_{cosine}X_3 + \beta_{polycentric}X_4 \quad (1)$$

$$Y = \beta_{linear}X_1 + \beta_{circular}X_2 + \beta_{cosine}X_3^2 + \beta_{polycentric}X_4^3 \quad (2)$$

### 2.2. Methodology

The proposed explainable geospatial machine learning (XGeoML) model integrates the spatial weighting principles from GWR with ML technologies to enhance the model's responsiveness to local spatial variability and predictive accuracy. By incorporating spatial weights into the model design, XGeoML adeptly captures the complexity and non-linearity inherent in spatial data (Fig. 2).

Fig. 2 outlines the workflow of the XGeoML model. The process begins with selecting an initial bandwidth and an appropriate spatial weighting scheme. This is followed by the construction and evaluation of local models for each point, utilizing leave-one-out (LOO) cross-validation for training, prediction, and evaluation. The next step involves iterating over different bandwidths to identify the optimal setting that maximizes model performance. Once the optimal bandwidth is determined, it is integrated with the machine learning model, and XAI tools such as SHAP, LIME, and Feature Importance are applied to achieve interpretable spatial variability coefficients. The final step includes partial dependency estimation, where percentile binning sampling on selected features is used to generate partial dependency estimations with the trained model.

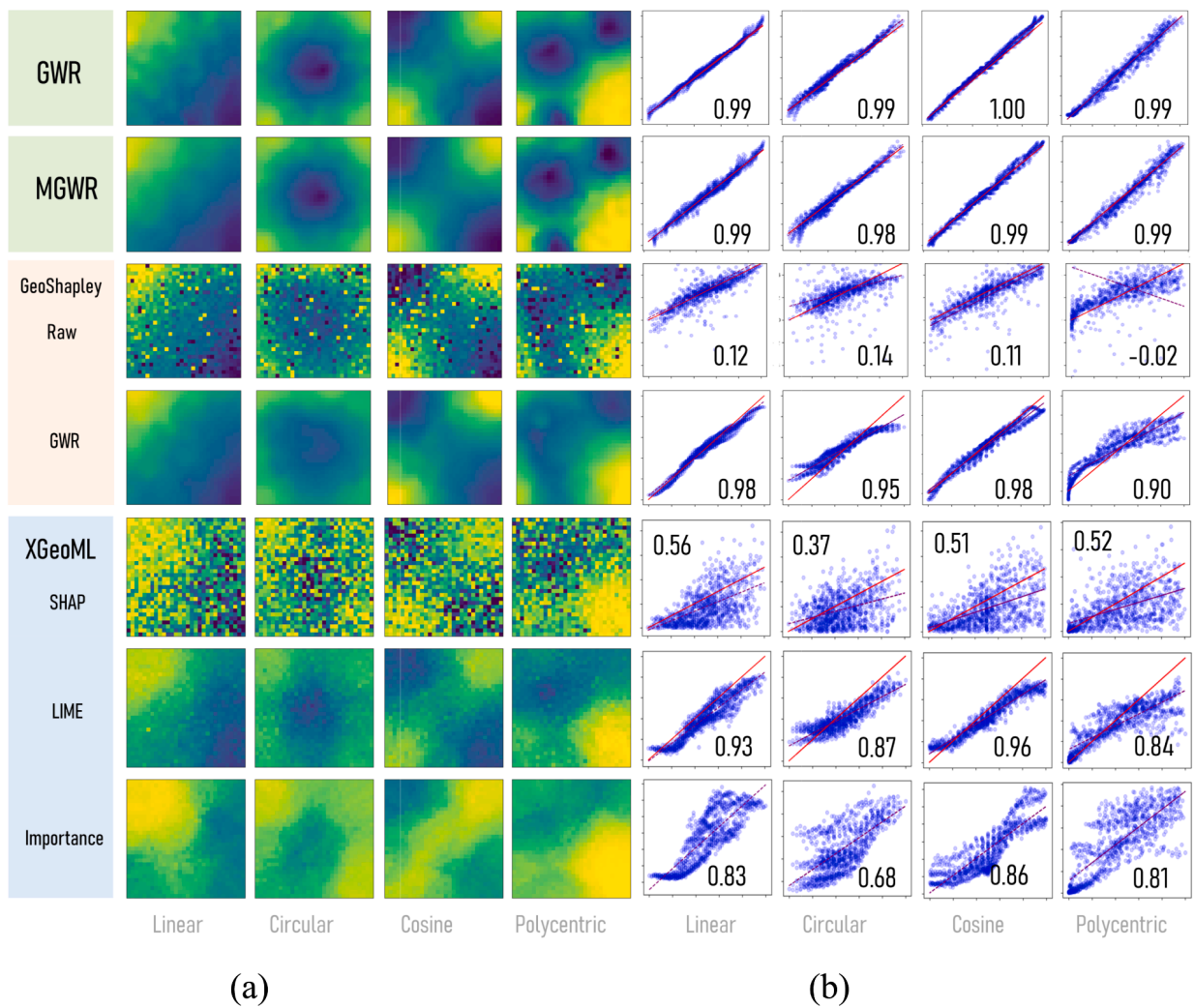

**Fig. 3.** Model comparison based on linear equations. (a) visualizes the spatially varing coefficients, (b) depicts the correlation between predicted coefficients and coefficients in the true model. Axes are scaled 0–5, barring GeoShapley's raw coefficients on the Y-axis, spanning −4 to 4, and XGeoML's Importance on the Y-axis, ranging 0–1. The red line, $y = x$, illustrates the ideal scenario where predicted values should equal the true values. Each column represents the target coefficients in the true model, with each row corresponding to different models. (For interpretation of the references to colour in this figure legend, the reader is referred to the web version of this article.)

**Table 1**
Model accuracy based on $R^2$.

| OLS | GWR | MGWR | GeoShapley* | | XGeoML- GBR-Fixed bandwidth | |
|---|---|---|---|---|---|---|
| | | | MLP | GBR | Binary | Gaussian Binary |
| 0.37 | 0.70 | 0.81 | 0.76 | 0.75 | 0.75 | 0.66 |

* MLP:Train 0.97, Test 0.51, GBR, Train: 0.96, Test:0.48.

### 2.2.1. Integration foundation for spatial weighting scheme and machine learning

The foundation of this XGeoML model is explained through mathematical formulations. In the framework of Ordinary Least Squares (OLS), the objective is to identify coefficients $\beta$ that minimize the sum of squared errors. The Geographically Weighted Regression (GWR) model, however, adjusts for each local point $i$ based on surrounding points and their spatial weights $w_i$, utilizing weighted least squares to depict spatial variability (Eq. (3)). This process effectively entails multiplying both explanatory and response variables by the square root of their spatial weights before conducting a standard OLS analysis. We propose that similarly processing the explanatory and response variables within a machine learning model—by multiplying them by the square root of their spatial weights—can more precisely model spatial variations, thereby improving the accuracy and stability of predictions.

$$S_W(\beta) = \sum_{i=1}^{n} w_i (y_i - x_i^T \beta)^2 ; \ S_W(\beta) = \sum_{i=1}^{n} \left( \sqrt{w_i} y_i - \sqrt{w_i} x_i^T \beta \right)^2 \quad (3)$$

### 2.2.2. Bandwidth and spatial weighting kernel selection

The choice of bandwidth and kernel significantly influences model performance in geospatial machine learning models. We explored three different kernel modes: Gaussian, Binary, and Gaussian Binary, each affecting the distribution of spatial weights and thus the model's predictive capacity and interpretability.

The Gaussian kernel utilizes a continuous Gaussian function to allocate spatial weights, with weights decreasing as distance increases, given by:

$$w_i = \exp\left(-\frac{d_i^2}{2\sigma^2}\right) \quad (4)$$

where $d_i$ represents the distance from the target point to other points, and $\sigma$ is the bandwidth parameter controlling the rate of spatial weight decay.

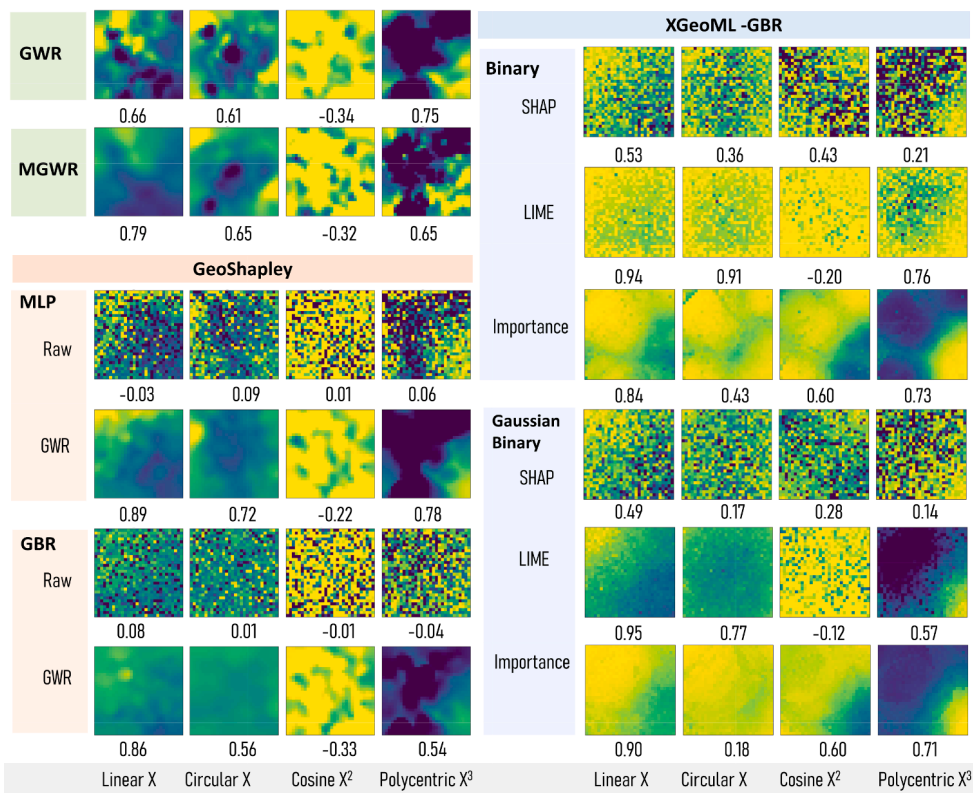

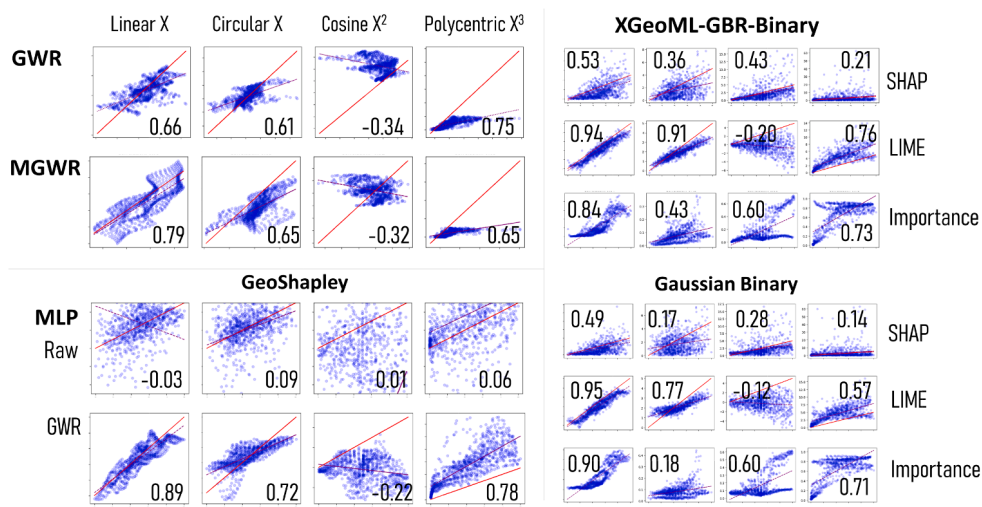

**Fig. 4.** Model comparison based on nonlinear equations. (a) visualizes the spatial variability coefficients, (b) depicts the correlation between predicted coefficients and coefficients in the true model, (c) show the GWR interpolated SHAP value of XGeoML. The red line, $y = x$, illustrates the ideal scenario where predicted values should equal the true values. Each column represents the target coefficients in the true model, with each row corresponding to different models. (For interpretation of the references to colour in this figure legend, the reader is referred to the web version of this article.)

The Binary kernel assigns a weight of 1 to points within the bandwidth range and 0 to points outside. This method is straightforward, ensuring that only points within a certain distance from the target point contribute to the prediction.

The Gaussian binary kernel combines the characteristics of Gaussian and Binary kernels, allocating weights according to a Gaussian distribution within the bandwidth and 0 outside.

*2.2.3. Local machine learning training and spatial interpretability*

Assuming that an independent machine learning model is constructed for each observation point simplifies the model training process, thereby avoiding complex hyperparameter optimization. In experiments, we utilized default parameter configurations of regression models from the *scikit-learn* Python packages and conducted model training and validation using leave-one-out (LOO) cross-validation.

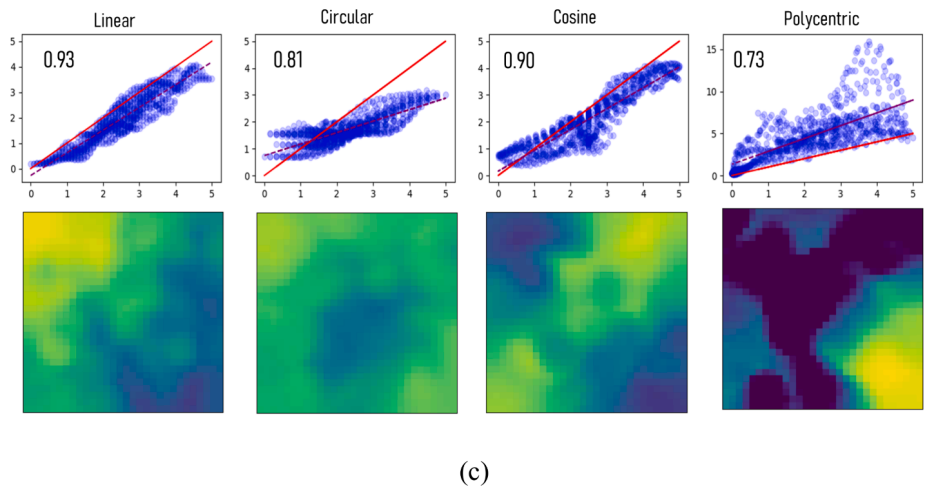

(c)

**Fig. 4.** (*continued*).

Interpretability is enhanced through the integration of explanatory XAI tools such as Local Interpretable Model-agnostic Explanations (LIME), SHapley Additive exPlanations (SHAP) for general models and Feature Importance in tree-based models.

LIME offers explanations by approximating the local decision boundary of complex models, with its formula represented as:

$$LIME(x) = \arg\min_{g \in G} L(f, g, \pi_x) + \Omega(g) \qquad (5)$$

where $f$ is the original model, $g$ is a simplified model (e.g., a linear model), $\pi_x$ is a function measuring the local neighborhood, $L$ is a loss function that measures how well $g$ approximates $f$ in the neighborhood of $x$, and $\Omega$ is a complexity penalty.

SHAP assigns a contribution value to each feature based on the Shapley value from game theory, with the formula:

$$\phi_j = \sum_{S \subseteq N\{j\}} \frac{|S|!(|N| - |S| - 1)!}{|N|} [f(S \cup \{j\}) - f(S)] \qquad (6)$$

where N is the set of all features, $S$ is a subset of features excluding feature j, $f(S)$ is the prediction of the model with feature set $S$, and $\phi_j$ is the Shapley value of feature j, indicating its average contribution to the model's prediction.

For tree-based models, such as random forests and gradient boosting trees, Feature Importance is commonly used to assess the importance of each feature in model predictions. Feature Importance can be calculated based on the frequency of feature usage in tree splits or the purity gain from the splits, with a common calculation method being:

$$Importance(x) = \sum_{t \in T} \Delta purity(t) \qquad (7)$$

where $T$ is the set of all tree nodes that split on feature $x$, and $\Delta purity(t)$ is the purity gain from splitting at node $t$.

For partial dependence estimation, a binning sampling method is employed to reduce the computational cost associated with large datasets. By dividing each feature variable into uniform intervals and either selecting sample points within each interval or generating uniformly distributed data, we efficiently estimate the model's partial dependence on various features with reduced computational cost.

### 2.2.4. Model comparison and evaluation

Using a synthetic dataset, we evaluated and compared the performance of various geospatial models, including GWR, MGWR, and GeoShapley. Our comparative analysis assessed the prediction accuracy ($R^2$) of different models, the correlation between spatial variability coefficients and the ground truth model, the performance of XGeoML model under various bandwidth forms and weighting methods, the variability in interpretability models across different bandwidths, and the performance of all machine learning models from the scikit-learn package using uniform parameter settings.

These comparisons examined the models' proficiency in capturing and elucidating spatial variability and their capability to discern the structure and dynamics inherent in spatial data. This holistic approach not only highlighted each model's strengths and weaknesses in predicting and understanding spatial phenomena but also shed light on the efficacy of XGeoML in leveraging spatial weighting principles and machine learning to enhance both predictive accuracy and model interpretability.

## 3. General comparison of linear model and nonlinear model

### 3.1. Linear model

Within the context of linear models of Eq. (1), all models show high overall fitting results $R^2$: OLS with 0.775, GWR with 0.989, GeoShapley with 0.956, and MGWR and XGeoML both with 0.992. GWR and MGWR accurately and stably capture the spatially varying coefficients of the explanatory variables in the synthetic data (Fig. 3). Conversely, the GeoShapley model, developed on the Multilayer Perceptron (MLP) neural network framework, exhibits deficiencies in accurately determining the original spatial variability coefficients, primarily due to the substantial presence of noise within the model. Despite this, most predictive values persistently align with the original spatial variability coefficients. Remarkably, after applying GWR for smoothing the initial dataset, GeoShapley's outcomes can emulate the spatial patterns exhibited by MGWR, enhancing predictive accuracy. This adaptation, however, may diminish the novel allure of employing GeoShapley.

The XGeoML model utilizes an adaptive bandwidth approach by selecting the nearest 150 data points, and integrates an ensemble model, Gradient Boosting Regressor, with three different explainability models. The results show that LIME outperforms in terms of accuracy, followed by Feature Importance, while SHAP coefficients are relatively lower. Overall, this indicates that XGeoML exhibits exceptional performance in processing linear models. Noteworthily, despite leveraging *sklearn* package's default parameter configurations, XGeoML secures commendable results, accentuating the pivotal contribution of spatial weighting towards augmenting model performance.

### 3.2. Nonlinear model

While applied to a nonlinear model based on Eq. (2), the $R^2$ values for all models decreased compared to the linear model, especially for OLS (Table 1). Wherein, the XGeoML model achieved an $R^2$ value of 0.95 on the training set and 0.75 on the test set. Although this performance is slightly below the 0.81 $R^2$ value of the MGWR model, it surpasses the GeoShapley model, which exhibited an impressive $R^2$ of 0.97 in the training set but experienced a substantial drop to an $R^2$ of only 0.50 in the test set. Moreover, when choosing a fixed bandwidth, the XGeoML's binary weight-based $R^2$ performance exceeded that of the Gaussian binary weights, highlighting the crucial impact of spatial weights on model performance.

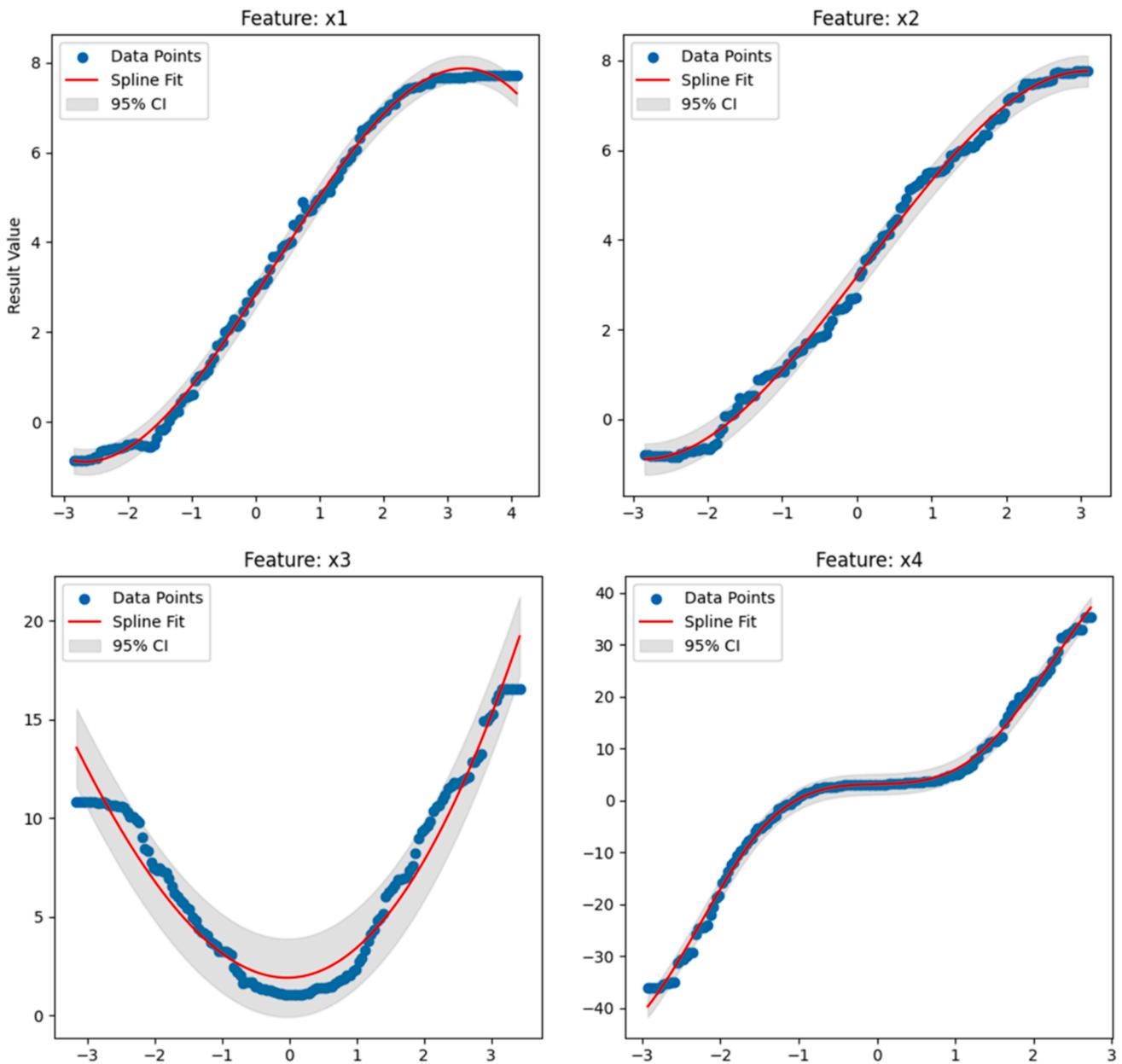

**Fig. 5.** Partial dependence plot of explanatory variables.

However, the visualization of spatial variability coefficients in Fig. 4 (a) reveals the XGeoML model's prominence. The result shows that GWR, MGWR, and GeoShapley models underperformed in capturing spatial variability coefficients related to the interaction between cosine spatial coefficients and the square of variables, whereas the XGeoML model effectively captured these changes. Although the GeoShapley model made some progress in capturing monotonic relationships (such as $x$ and $x^3$) through optimization of the GWR model, it did not exhibit competitive performance in capturing the interaction between cosine spatial variability features and the square of variables, whether based on MLP or GBR. This outcome emphasizes the challenges models face when dealing with complex spatial changes and non-monotonic variable interactions.

The scatter plot for correlation in Fig. 4(b) verifies the interpretative capability provided by the explainers of SHAP and Feature Importance which showed positive correlations in both binary and Gaussian binary weighting method. Notably, the LIME analysis outperformed GWR, MGWR, and GeoShapley on three other coefficients based on monotonous relationships. Although SHAP showed positive correlation relative to LIME, its weaker influence suggests potential for improvement in nonlinear model performance and possibly explains GeoShapley's interpretability shortcomings. The XGeoML model's Feature Importance analysis was particularly effective in capturing changes in cosine spatial variability and the square of variables $x$. This highlights the XGeoML model's comprehensive advantage in interpretability and accuracy under diverse and complex conditions. It is worth noting that when GWR is applied to perform interpolation analysis on the SHAP values of the XGeoML model, its interpretability is greatly enhanced, similar to GeoShapley, with higher accuracy in coefficient interpretation (Fig. 4c).

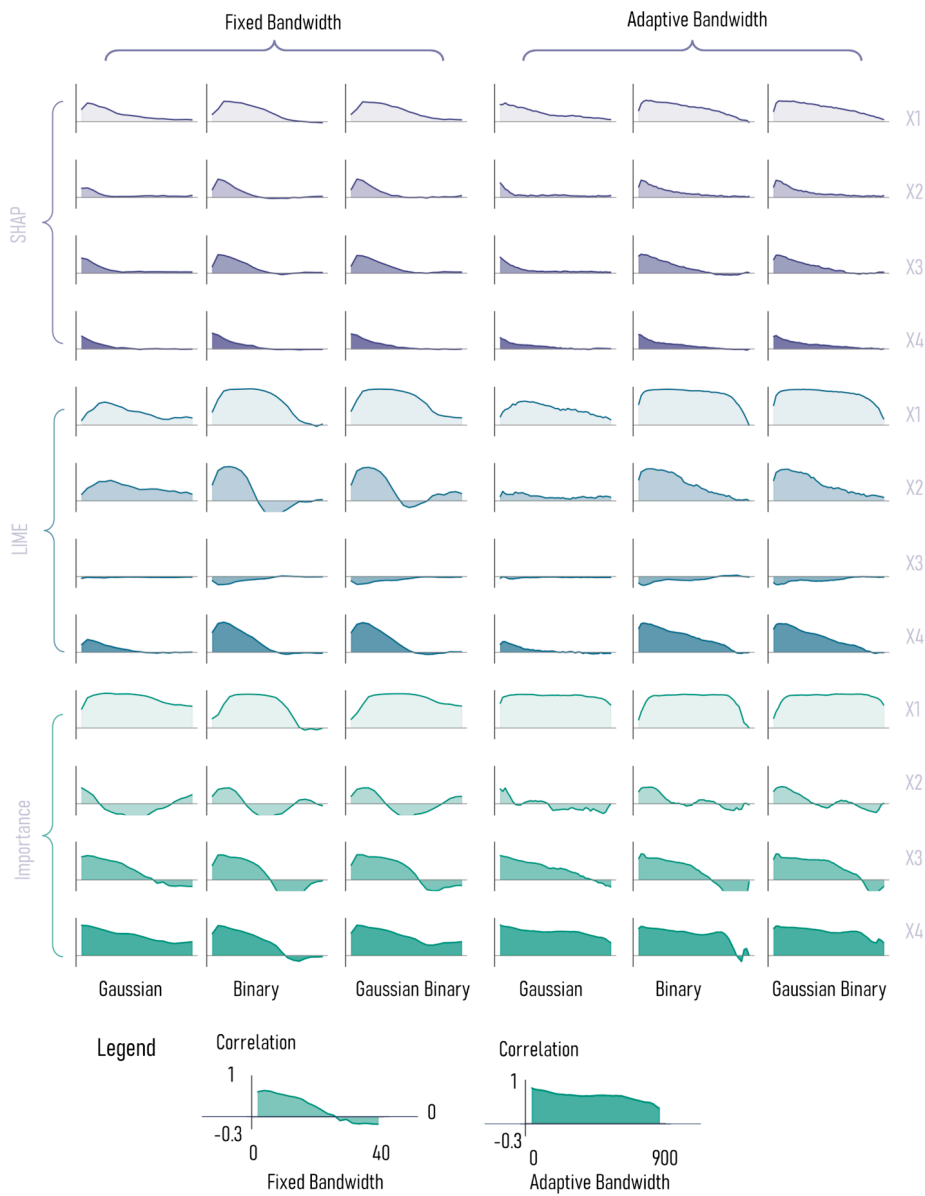

**Fig. 6.** Model performance with different bandwidths and spatial weighting kernels. Each row illustrates how the correlation between the estimated spatial variability coefficients by the model and those of the actual model shifts with changing bandwidth. The X-axis specifies the bandwidth, while the Y-axis measures the correlation, spanning from −0.3 to 1. The graph includes a reference line at y = 0 to aid in visual comparison. For bandwidth settings on the X-axis, fixed bandwidth ranges from 2 to 40, indicating a more localized consideration of spatial data, whereas adaptive bandwidth extends from 20 to 900.

The results of partial dependency estimations are presented in Fig. 5. Both GeoShapley and XGeoML accurately captured the variation patterns of the four explanatory variables, particularly the quadratic function of $x^3$ and the cubic function of $x^4$. It is important to note that GWR and MGWR do not provide partial dependency estimations, limiting their ability to analyze the variation trends of these nonlinear variables. Consequently, GWR and MGWR fall short in offering a comprehensive understanding of how explanatory variables interact and vary, particularly in nonlinear contexts.

## 4. Comparison of spatial weighting scheme and ML models in XGeoML

### 4.1. Bandwidth type and spatial weights kernel comparison

The evaluation of the accuracy of spatially varying coefficients, conducted using the GBR model across different bandwidths and spatial weight types, reveals consistent trends. Fig. 6 illustrates that the correlation of SHAP values remains positive across bandwidths but exhibits a discernible threshold decline as bandwidth increases. The influence of selecting different bandwidth types and weight methods on SHAP's spatial variability coefficient estimations appears minimal. In contrast, LIME's performance benefits significantly from adaptive bandwidth compared to fixed bandwidth, suggesting an insensitivity to the choice of bandwidth.

The Feature Importance metric, notable for its higher accuracy in previous comparison, warrants detailed examination. The analysis across various bandwidth and spatial weight combinations reveals that the bandwidth size selection does not markedly affect the $x_1$ variable, suggesting that linear spatial changes represent a global variable phenomenon. This pattern is similarly observed with the $x_4$ variable. However, the Cosine spatial effect on the $x_3$ variable shifts to negative beyond a certain bandwidth threshold, classifying it as a coefficient of mid-spatial scale variability. Among the different weighting methods,

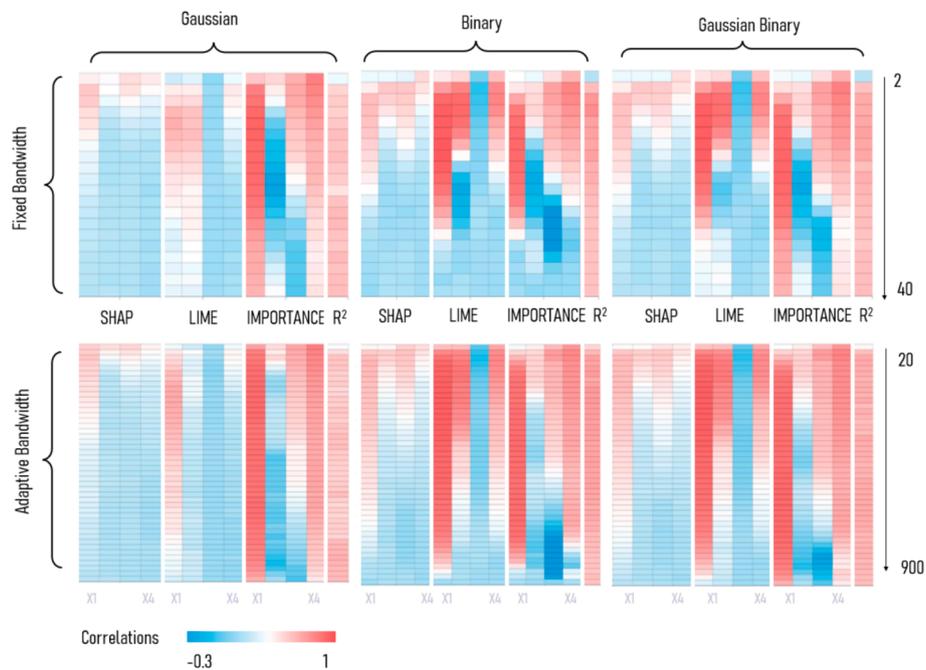

**Fig. 7.** Correlation Heatmap across various bandwidth types and spatial weighting schemes. This heatmap is organized into six panels based on a combination of three spatial weighting methods—Gaussian, Binary, and Gaussian Binary—and two bandwidth types: Adaptive and Fixed. Each panel presents the correlation values for explanatory variables, utilizing SHAP, LIME, and Feature Importance, with the final column dedicated to the $R^2$ values.

Gaussian weighting kernels exhibit the poorest performance, whereas Binary and Gaussian binary weights demonstrate more stability. This stability is especially apparent in the variability graph of the $x_4$ variable under the Importance metric. The heatmap in Fig. 7 further underscores that adaptive bandwidth provides a more robust adaptability to bandwidth changes, indicating that the Gaussian binary choice may offer a more stable option in XGeoML models.

*4.2. Optimizing model selection by balancing performance, efficiency, and accuracy*

This session evaluated and compared the efficacy of 28 regression models with the same Adaptive Binary spatial weighting scheme, encompassing ensemble techniques (such as Extra Trees and Random Forest), tree-based models, Nearest Neighbors methods, Neural Networks, SVMs, and linear models. Fig. 8 illustrates the performance of these models under varying bandwidth conditions, while Table 2 compiles their evaluation based on essential metrics like the coefficient of determination ($R^2$), execution time, and average correlation. The results revealed that the Extra Trees Regressor emerged as the top-performing model in terms of data fitting, achieving an $R^2$ value of 0.765 with a bandwidth of 180. However, its comparatively longer execution time highlights the crucial need for a trade-off between performance and computational efficiency during model selection.

While ensemble methods like Extra Trees and Random Forest showed superior performance, their longer runtime may not be suitable for applications requiring fast response times. Conversely, although linear models did not achieve the highest $R^2$ values, their consistent performance and shorter runtime make them a reliable choice for situations demanding quick model responses. Notably, the Gaussian Process performed inadequately on our dataset (with an $R^2$ value close to 0), indicating a possible mismatch with the data characteristics or a need for more precise parameter tuning. Similarly, the performance of Support Vector Machines (SVM) in this study was subpar, especially in terms of $R^2$ values, suggesting that these models might not be the best option for handling such data.

Furthermore, we observed a relationship between the average correlation of spatially varying coefficients and model accuracy. Models with higher $R^2$ values not only demonstrated excellent fitting capabilities but also showed higher average correlations, indicating a strong relationship between predicted and actual values. However, some models, like the Histogram Gradient Boosting, despite having commendable $R^2$ values, exhibited very low average correlations for coefficients, revealing that a high $R^2$ value in prediction does not necessarily imply better accuracy in spatial varying coefficients, emphasizing the importance of considering multiple metrics when evaluating spatial model performance.

**5. Discussion and conclusion**

The ensemble framework for Explainable Geospatial Machine Learning (XGeoML) models is proposed to tackle critical challenges in geographic analysis, particularly the complexity and non-linearity inherent in spatial data. By integrating local spatial weighting methods from GWR with advanced ML techniques and XAI tools, XGeoML offers an integrated approach to capturing and interpreting spatial variability.

Comparative analysis reveals that XGeoML models accurately capture the spatial variation patterns of explanatory variables, notably the interaction of quadratic function and cosine spatial effect, while GWR, MGWR and GeoShapley fail to do so. This demonstrates that local spatial weighting can be extended to general machine learning models (Fotheringham et al., 2017). The result also indicates that while models like GWR and MGWR may show high predictive accuracy, their ability to estimate spatial coefficients accurately can be poor while handling non-linearity. This discrepancy underscores the need for using multiple models to cross-validate the result and better understand the performance (Li, 2016).

The XAI interpreters like SHAP, LIME, and Feature Importance integrated in XGeoML not only enhance interpretability but also provides deeper insights into the spatially varying effects. Generally, Feature Importance offered by tree models can provide stable explanations for spatially varying effects (Grekousis, 2022). Similar to the challenges faced by GeoShapley (Li, 2022), the noise of SHAP within XGeoML can also prevents accurate capturing of these effects. However, GWR-interpolated SHAP can indeed provide more accurate spatially varying coefficients. Moreover, the integration of partial dependency estimation methods in XGeoML aids in explaining the nonlinear relationships between explanatory variables and explained variables within a geographic context.

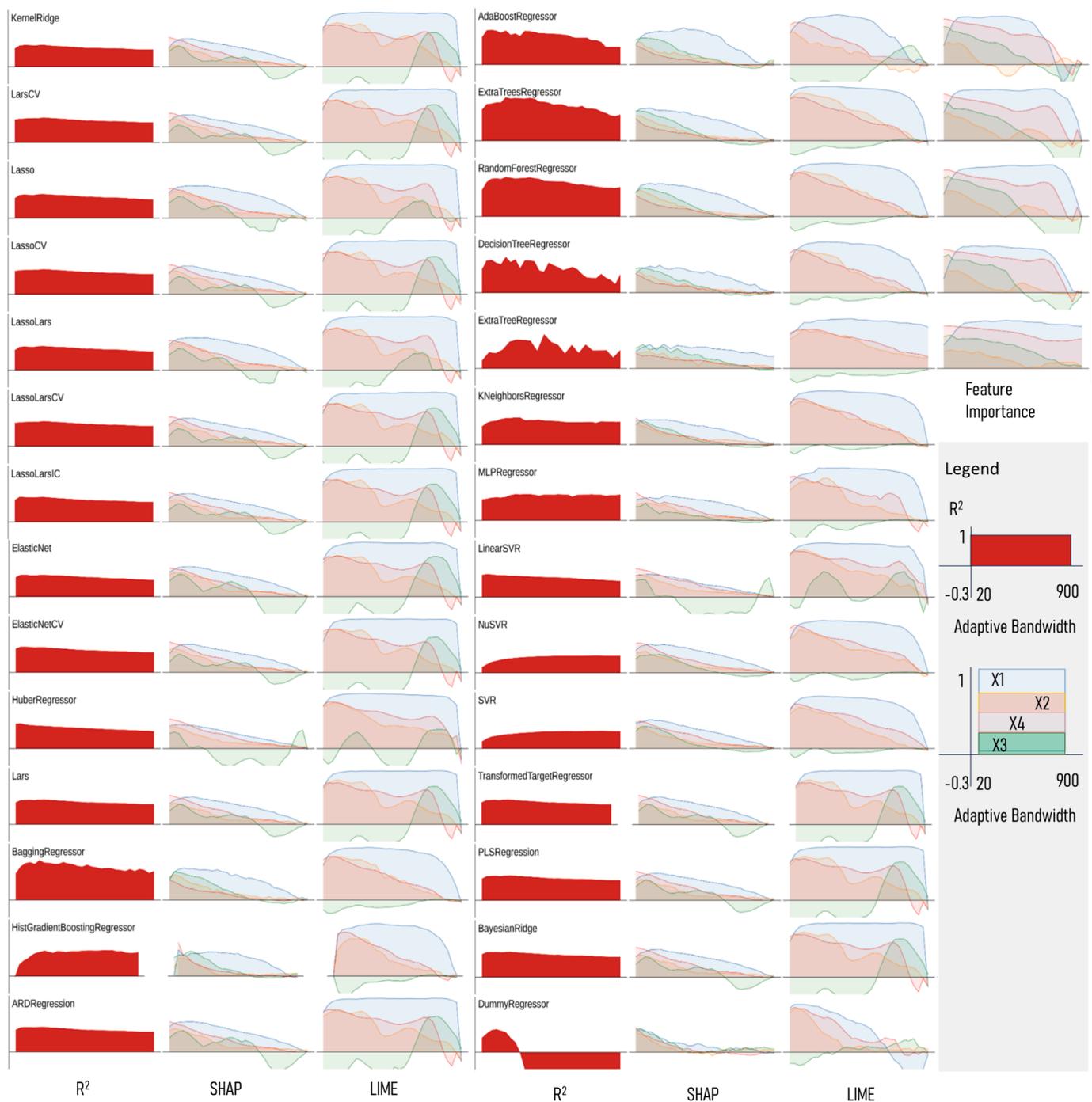

**Fig. 8.** 28 XGeoML models' performance with the changing bandwidth. The X-axis specifies the bandwidth, while the Y-axis measures the correlation, spanning from −0.3 to 1. The graph includes a reference line at y = 0 to aid in visual comparison. For bandwidth settings on the X-axis, the adaptive bandwidth extends from 20 to 900.

Comparing the prediction accuracy, the correlation between spatially varying coefficients, the performance of XGeoML under different bandwidth types, spatial weighting methods and ML models reveals that it is crucial to compare multiple ML models and their parameters. This necessity can pose challenges for model computation, reproducibility and uncertainty (Kedron et al., 2023).

Despite these advancements, the XGeoML model presents areas for improvement and opens avenues for future research. One limitation lies in the model's computational efficiency, particularly when handling large datasets or requiring real-time analysis. The balance between model complexity and computational demand necessitates optimization to ensure broader applicability and faster processing times. Moreover, the challenge of selecting optimal bandwidth and kernel types remains. The impact of these choices on model performance underscores the need for more adaptive and data-driven methods to determine these parameters, potentially through automated tuning processes or ML algorithms.

Table 2
XGeoML Model performance comparison.

| Model name | Model type | Runtime (s) | R2 | Average Correlation | Optimal Bandwidth |
| --- | --- | --- | --- | --- | --- |
| ExtraTreesRegressor | Ensemble | 6043 | 0.765 | 0.631 | 180 |
| BaggingRegressor | Ensemble | 1498 | 0.707 | 0.366 | 180 |
| RandomForestRegressor | Ensemble | 8720 | 0.701 | 0.637 | 180 |
| DecisionTreeRegressor | Tree | 386 | 0.630 | 0.595 | 180 |
| AdaBoostRegressor | Ensemble | 3244 | 0.616 | 0.619 | 90 |
| ExtraTreeRegressor | Tree | 348 | 0.607 | 0.523 | 300 |
| KNeighborsRegressor | Neighbors | 972 | 0.476 | 0.255 | 270 |
| MLPRegressor | Neural Network | 8788 | 0.466 | 0.239 | 300 |
| HistGradientBoostingRegressor | Ensemble | 5690 | 0.456 | 0.072 | 660 |
| ElasticNetCV | Linear | 1733 | 0.454 | 0.427 | 90 |
| LassoLarsIC | Linear | 347 | 0.446 | 0.413 | 60 |
| BayesianRidge | Linear | 386 | 0.445 | 0.317 | 210 |
| HuberRegressor | Linear | 425 | 0.443 | 0.400 | 60 |
| LarsCV | Linear | 509 | 0.443 | 0.310 | 210 |
| LassoLarsCV | Linear | 455 | 0.443 | 0.310 | 210 |
| LassoCV | Linear | 1516 | 0.443 | 0.312 | 210 |
| Lars | Linear | 351 | 0.442 | 0.316 | 210 |
| TransformedTargetRegressor | Compose | 313 | 0.442 | 0.316 | 210 |
| ARDRegression | Linear | 441 | 0.442 | 0.336 | 210 |
| PLSRegression | Cross_Decomposition | 316 | 0.442 | 0.316 | 210 |
| Lasso | Linear | 321 | 0.426 | 0.390 | 210 |
| LassoLars | Linear | 317 | 0.426 | 0.390 | 210 |
| DummyRegressor | Dummy | 3364 | 0.408 | 0.292 | 120 |
| LinearSVR | SVM | 317 | 0.404 | 0.394 | 60 |
| ElasticNet | Linear | 409 | 0.385 | 0.453 | 90 |
| KernelRidge | Kernel Ridge | 549 | 0.384 | 0.306 | 210 |
| SVR | SVM | 5298 | 0.308 | 0.036 | 780 |
| NuSVR | SVM | 3164 | 0.306 | 0.053 | 750 |
| GaussianProcessRegressor | Gaussian Process | 236 | 0.005 | 0.168 | 180 |

**CRediT authorship contribution statement**

**Lingbo Liu:** Writing – review & editing, Writing – original draft, Software, Methodology, Conceptualization.

**Declaration of competing interest**

The authors declare that they have no known competing financial interests or personal relationships that could have appeared to influence the work reported in this paper.

**Data availability**

The example data and Google Colab notebook are available at https://github.com/UrbanGISer/XGeoML. The Python Package XGeoML can be installed via PYPI, https://pypi.org/project/XGeoML

**Acknowledgement**

This work is partially funded by NSF grant #1841403, the National Natural Science Foundation of China, #51978535 and #52078390.